\documentclass[letterpaper, 10 pt, conference]{ieeeconf}
\IEEEoverridecommandlockouts
\usepackage{amsmath,amssymb}
\usepackage{bm}
\usepackage{algorithm}
\usepackage{algpseudocode}
\usepackage{multirow}
\usepackage{graphicx}
\usepackage{xcolor}
\usepackage{url}
\usepackage{booktabs}
\usepackage[table]{xcolor}
\usepackage{comment}
\usepackage[colorlinks=true,allcolors=blue]{hyperref}

\usepackage{amsthm}

\theoremstyle{definition}
\newtheorem{proposition}{Proposition}
\newtheorem{definition}{Definition}
\newtheorem{remark}{Remark}

\title{\LARGE \bf{DFLAT: Model-based Diffusion on Differentially Flat Spaces for Long-Horizon Trajectory Optimization}}
\author{Ege Yuceel$^{1}$, Taekyung Kim$^{2}$, Keyvan Majd$^{3}$, Hideki Okamoto$^{3}$,\\
Bardh Hoxha$^{3}$, Sayan Mitra$^{1}$, Georgios Fainekos$^{3}$%
\thanks{This research was performed while Ege Yuceel and Taekyung Kim were interns at Toyota Motor North America, Research \& Development.}
\thanks{$^{1}$Department of Electrical and Computer Engineering, University of Illinois Urbana-Champaign, Urbana, IL, 61801, USA
        {\tt\small \{eyceel2, mitras\}@illinois.edu}}%
\thanks{$^{2}$Department of Robotics, University of Michigan, Ann Arbor, MI, 48109, USA
        {\tt\small taekyung@umich.edu}}%
\thanks{$^{3}$Toyota Motor North America, Research \& Development, Ann Arbor, MI, 48105, USA
        {\tt\small <first\_name.last\_name>@toyota.com}}%
}

\begin{document}

\maketitle

\begin{abstract} Long-horizon diffusion-based planning for kinodynamically constrained systems requires both scalability and feasibility, yet existing diffusion planners typically sacrifice one for the other. We present \textbf{DFLAT}, a diffusion-based planning algorithm that operates directly on a differentially flat trajectory representation. Instead of diffusing over time-discretized controls, DFLAT denoises a compact set of Hermite curve parameters in the flat-output space, consisting of intermediate waypoints together with the endpoint tangents of each segment. Every denoised sample maps to a dynamically consistent trajectory, so feasibility holds by construction rather than being restored afterward. Experiments on challenging tractor--trailer planning benchmarks show that DFLAT matches or exceeds the success rates of existing baselines at consistently
lower planning times, particularly on long-horizon problems requiring multiple forward--reverse maneuvers. In cases where kinodynamic RRT requires over 100 seconds, DFLAT plans in subsecond time, enabling practical deployment. \href{https://yuceelege.github.io/dflat}{\textcolor{red}{\textbf{[Project Page]}}}%
\footnote{Project webpage: \href{https://yuceelege.github.io/dflat}{\textcolor{black}{\nolinkurl{https://yuceelege.github.io/dflat}}}}
\end{abstract}

\section{Introduction}

Planning motions that are both collision-free and dynamically feasible is a long-standing challenge in robotics, particularly for nonholonomic systems operating over long horizons \cite{lavalle2001randomized}. Recent advances in generative modeling, including diffusion models and flow matching, have introduced powerful new approaches to robot policy learning and motion planning~\cite{urain2024deep}. Diffusion-based planners, in particular, have demonstrated strong performance across a broad range of robotic planning tasks~\cite{ho2020denoising, chen2024simple}.

These results are harder to sustain once the dynamics are constrained and the horizon is long. Every candidate trajectory must remain
dynamically consistent, yet the sample budget must not grow with the
horizon, and existing representations trade one against the other. One
route to feasibility is to diffuse over the control sequence rather than
the state trajectory: samples are drawn from a distribution favoring
low-cost controls and rolled forward through the known dynamics, so the
resulting trajectories are dynamically feasible by
construction~\cite{pan2024model, williams2015model}. This representation
scales poorly with the horizon: low-cost controls occupy a shrinking
fraction of the sample space, so the sample budget must grow
accordingly. An alternative is to diffuse over the
trajectory directly~\cite{li2023hierarchical, chen2024simple,
srikanth2025gpd}, which avoids this conditioning problem, as obstacle and
goal constraints act on the sampled variables themselves rather than
through a long composition of dynamics steps. The resulting trajectory is
no longer the rollout of any admissible control and is therefore not
necessarily dynamically feasible, requiring projection or post-hoc
refinement to recover a trackable solution~\cite{christopher2024constrained,
christopher2025neuro, romer2025diffusion, kondo2024cgd}. To the best of our knowledge, no existing diffusion-based planner is both feasible by construction and efficient as the horizon grows for such systems.

We resolve this tension through the choice of trajectory representation.
A particular class of systems, called differentially flat, admits a set
of outputs from which the states and inputs can be
recovered~\cite{fliess1995flatness}. In the flat-output space, a
sufficiently smooth curve is therefore already a valid state and input
trajectory, so feasibility follows from the representation rather than
being restored after the fact. Our method diffuses over a compact
parametrization of such a curve: a sequence of waypoints together with
the in-segment parameters that join them into a smooth path. Because
these parameters describe segments rather than time steps, the design
space scales with the number of segments and the structure of the
maneuver, not with the resolution of the discretization, and stays small
as the horizon grows. Our method thus retains the dynamic consistency of control-sequence diffusion without the large sample budget it requires.

\begin{figure*}[t]
    \centering
    \includegraphics[width=1.\textwidth]{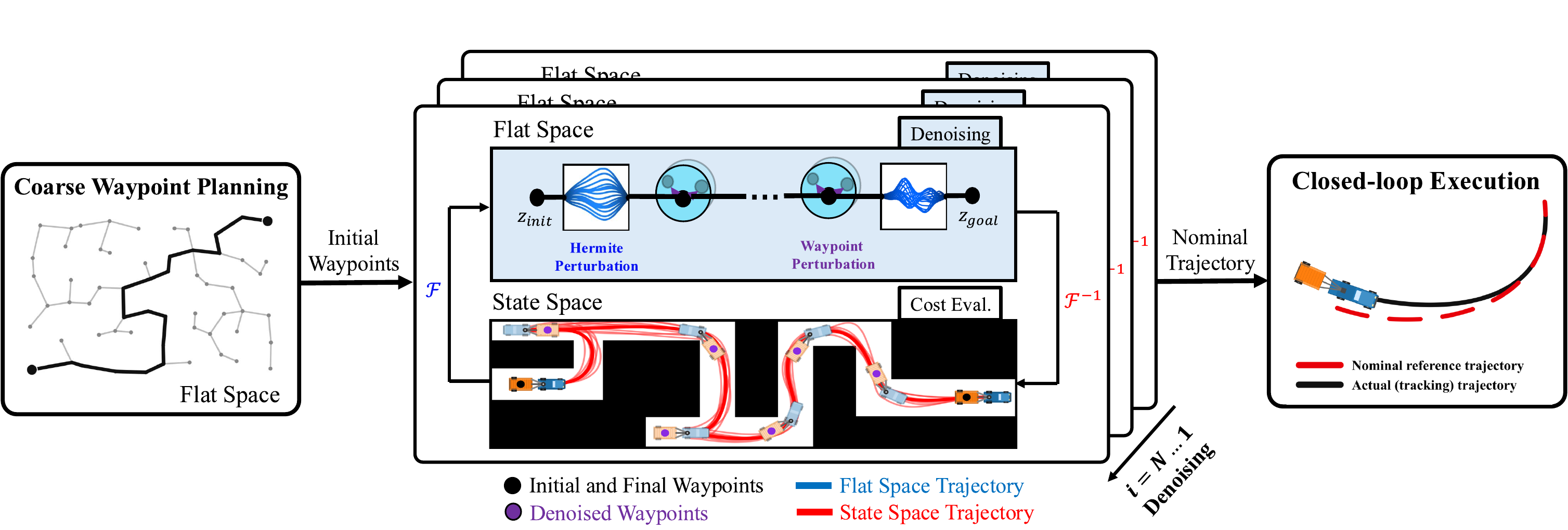}
    \caption{Overview of DFLAT. Given a long-horizon planning task requiring multiple
forward and reverse motions (bottom), Coarse Waypoint Planning (CWP) produces a
coarse waypoint prior in flat-output space. At each reverse diffusion step
$i = N, \ldots, 1$, candidates are drawn by perturbing the waypoint poses and
the segment-local Hermite parameters, lifted through the inverse flatness map
into dynamically consistent trajectories, and scored under the planning cost.
The score-weighted mean seeds the next step, refining the coarse prior into a
low-cost trajectory for the downstream tracking controller.}
    \label{fig:main_figure}
\end{figure*}

\subsection{Related Work}

Model-based diffusion (MBD) uses the dynamics model to compute a training-free score function for trajectory optimization~\cite{pan2024model},
and Safe Model Predictive Diffusion (Safe MPD) combines this with shielded rollouts to guarantee safety
throughout sampling process~\cite{kim2026safempd}. Barrier-rate guided MPPI
embeds barrier constraints into the sampling distribution for tractor--trailer
maneuvers in narrow spaces~\cite{majd2025gpu}. These methods achieve kinodynamic
feasibility by construction, but they operate on representations whose
dimension grows with the planning horizon, causing sample efficiency to
degrade on long-horizon problems. Recent work has also explored more
compact parameterizations for diffusion planning. Guided Polynomial Diffusion (GPD) performs diffusion directly over Bernstein
polynomial trajectory coefficients~\cite{srikanth2025gpd}, and sparse control
points with spline interpolation have been used to obtain smooth control
trajectories from few samples in MPPI~\cite{miura2024spline}. Hierarchical
diffusion methods instead plan over temporally abstracted
sub-goals~\cite{li2023hierarchical,chen2024simple}, and coupled hierarchical
diffusion tightens the interaction between the two levels for long-horizon
tasks~\cite{hao2025chd}. Although these approaches
improve horizon scalability, the sampled coefficients and sub-goals need
not correspond to admissible motions. Feasibility is instead recovered afterward, for instance by filtering out
samples that violate the constraints~\cite{yun2024guided}, or by projecting the
sampled variables onto the constraint
set~\cite{christopher2024constrained,romer2025diffusion}.
A representative instance related to our work diffuses over control points
rather than over the trajectory itself, and recovers dynamic feasibility with a
quadratic program~\cite{kondo2024cgd}, which is convex only when the relevant
bounds are affine in the sampled variables and does not extend to nonholonomic
models.

A different way to obtain feasibility is through the choice of coordinates. Differential flatness provides a change of variables under which
trajectories in the flat-output space map back to recoverable state-input
trajectories~\cite{fliess1995flatness}. Beyond recoverability, flatness
provides a compact trajectory representation that has been widely
exploited for motion planning and trajectory
generation~\cite{rouchon1993flatness}. Existing
planners operating directly in the flat-output space are largely
sampling-based. Flat-RRT*~\cite{bascetta2017flat} precomputes motion
primitives to expand the search tree in the flat-output space, but
planning performance depends strongly on the primitive library, with finer
discretizations increasing both memory requirements and computation time. The
same dependence appears in DE-AGT~\cite{zheng2025delayed}, which pairs a
precomputed primitive library with a learned cost-to-go heuristic for
tractor--trailer parking. More recently,~\cite{duong2026ultrafast} connects sampled states with
closed-form segments, at the cost of restricting the objective to costs
quadratic in the control.
All of these operate by search. We instead pose planning in the flat-output space as diffusion based trajectory optimization, which places no restriction on the cost and
requires no primitive library.

In this paper, we propose \textbf{DFLAT}, a diffusion-based trajectory
optimization algorithm for long-horizon kinodynamic planning. The main
contributions of this work are:

\begin{itemize}
\item We formulate model-based diffusion directly in the differentially flat
output space, shifting the optimization problem from time-discretized control
sequences to a low-dimensional trajectory parameterization.
\item We introduce a segmented Hermite representation that guarantees every
denoised sample corresponds to a dynamically consistent trajectory.
\item We validate DFLAT on long-horizon tractor--trailer planning, where it
solves every benchmark instance orders of magnitude faster than
baselines, reaching subsecond latency on short parking scenarios.
\item We release a benchmark library of long-horizon tractor--trailer planning
problems together with our implementation.
\end{itemize}

\section{Preliminaries}
\subsection{Problem Formulation}
Let $\mathcal{M} \subset \mathbb{R}^d$ denote a known map of the environment. Consider a continuous-time system
\begin{equation} \label{eq:system}
    \dot{\bm{x}}(t) = f(\bm{x}(t),\bm{u}(t)), \qquad t \in [0,T],
\end{equation}
with state $\bm{x}(t) \in \mathcal{X} \subset \mathbb{R}^n$, control $\bm{u}(t) \in \mathcal{U} \subset \mathbb{R}^m$, and variable final time $T$. Let $
\tau_{\bm{x}} := \bm{x}(\cdot), \ \tau_{\bm{u}} := \bm{u}(\cdot)$
denote the state and control trajectories over $[0,T]$, and let
$
    Y := (\tau_x,\tau_u)
$
denote the full trajectory. We consider the following trajectory optimization problem: given a start state $\bm{x}_{\text{init}}$ and goal state $\bm{x}_{\text{goal}}$,
\begin{subequations}
\begin{align}
    Y^* := \operatorname*{arg\,min}_{\tau_{\bm{x}},\tau_{\bm{u}}} \quad & J(Y) \\
    \text{s.t.} \quad
    & \dot{\bm{x}}(t) = f(\bm{x}(t),\bm{u}(t)), \quad t\in[0,T], \\
    & \bm{x}(0) = \bm{x}_{\text{init}}, \quad \bm{x}(T) = \bm{x}_{\text{goal}}, \\
    & g(\bm{x}(t)) \ge 0,\quad t \in [0,T], \\
    & \bm{u}(t) \in \mathcal{U}, \quad t \in [0,T],
\end{align}
\end{subequations}
where $J$ is the cost relevant to task, and $g : \mathcal{X} \to \mathbb{R}$ represents the constraint function induced by the map $\mathcal{M}$ together with any additional dynamical constraints.
\subsection{Model-based Diffusion}
\label{sec:mbd}
Model-based diffusion~\cite{pan2024model} formulates trajectory optimization as sampling
from a distribution that assigns higher probability to lower-cost
trajectories. Given the trajectory cost $J(Y)$, define
\begin{equation}
    p_0(Y) \propto \exp\left(-\frac{J(Y)}{\lambda}\right),
    \qquad \lambda > 0,
\end{equation}
where $\lambda$ is a temperature parameter, so that smaller values of
$\lambda$ concentrate the distribution around lower-cost trajectories.
Diffusion models sample from this distribution by starting from noise and
iteratively denoising~\cite{ho2020denoising}. Let $\{\beta_i\}_{i=1}^{M} \subset (0,1)$ be an increasing noise
schedule, with $\alpha_i := 1-\beta_i$ and $\overline{\alpha}_i := \prod_{k=1}^{i}\alpha_k$, so that $\overline{\alpha}_i$ decreases from near one to near zero. The forward noising process satisfies
\begin{equation}
    Y^{(i)}
    =
    \sqrt{\overline{\alpha}_i}Y^{(0)}
    +
    \sqrt{1-\overline{\alpha}_i}\epsilon,
    \qquad
    \epsilon \sim \mathcal{N}(0,I).
\end{equation}
The reverse process initializes $Y^{(M)} \sim \mathcal{N}(0,I)$ and
updates the trajectory by
\begin{equation} \label{eq:reversestep}
    Y^{(i-1)}
    =
    \frac{1}{\sqrt{\alpha_i}}
    \left(
        Y^{(i)}
        +
        (1-\overline{\alpha}_i)
        \nabla_{Y^{(i)}} \log p_i(Y^{(i)})
    \right).
\end{equation}
Unlike learned diffusion models, model-based diffusion updates the trajectory
by Monte Carlo score ascent on $\nabla_{Y^{(i)}} \log p_i(Y^{(i)})$, using
sampled trajectories and the known objective. At step $i$, draw $K$ candidate
trajectories
\begin{equation}
    Y_k^{(i)}
    \sim
    \mathcal{N}
    \left(
        \frac{Y^{(i)}}{\sqrt{\overline{\alpha}_i}},
        \left(\frac{1}{\overline{\alpha}_i}-1\right)I
    \right),
    \qquad k=1,\ldots,K.
\end{equation}
The schedule therefore sets the sampling radius, which shrinks as $i$ decreases. These samples are combined through the cost-weighted average
\begin{equation} \label{eq:softmax}
    \overline{Y}^{(i)}
    =
    \frac{
        \sum_{k=1}^{K} Y_k^{(i)}
        \exp\left(-J(Y_k^{(i)})/\lambda\right)
    }{
        \sum_{k=1}^{K}
        \exp\left(-J(Y_k^{(i)})/\lambda\right)
    }.
\end{equation}
Then, the score function at step $i$ is approximated as
\begin{equation} \label{eq:estimatescore}
    \nabla_{Y^{(i)}} \log p_i(Y^{(i)})
    \approx
    -
    \frac{Y^{(i)}}{1-\overline{\alpha}_i}
    +
    \frac{\sqrt{\overline{\alpha}_i}}{1-\overline{\alpha}_i}
    \overline{Y}^{(i)}.
\end{equation}
This yields a training-free score estimate that requires only evaluating the cost on sampled candidates.
\subsection{Differential Flatness}

Trajectory generation for a nonholonomic system normally means choosing a
control sequence and rolling the dynamics forward to obtain the resulting
states. For a particular class of systems, known as differentially flat,
this is unnecessary. The state and input can be recovered algebraically
from a set of flat-outputs and finitely many of their
derivatives~\cite{fliess1995flatness}, without integrating the dynamics.

\begin{definition}[Differential Flatness]
\label{def:flatness}
The control system \eqref{eq:system} is \emph{differentially flat} if there exists an output
$\bm{z}\in\mathbb{R}^{m}$, called the \emph{flat-output},
such that the state and input can be expressed algebraically in terms of
$\bm{z}$ and a finite number of its derivatives:
\begin{equation}
    \bm{x} = \Phi\bigl(\bm{z},\dot{\bm{z}},\ldots,\bm{z}^{(r)}\bigr),
    \qquad
    \bm{u} = \Psi\bigl(\bm{z},\dot{\bm{z}},\ldots,\bm{z}^{(r+1)}\bigr).
    \label{eq:recovery}
\end{equation}
where we refer to $\Phi$ and $\Psi$ as the \emph{state and input recovery maps}, respectively.

\end{definition}
Thus a sufficiently smooth choice of $\bm{z}(t)$ determines a recoverable
state-input trajectory $(\bm{x}(t),\bm{u}(t))$. This recasts trajectory generation as an unconstrained problem in the flat-output space, with dynamic consistency enforced structurally rather than as an explicit constraint.

\begin{remark}
For the car with $n$ trailers, the flat-output is the planar position of the
last trailer's axle midpoint, a vector with $m = 2$ components, and the
flat-output derivatives associated with a waypoint pose follow in closed form
from the Frenet-frame relations~\cite{rouchon1993flatness}.
\end{remark}

\subsection{Hermite Path Generation}
After choosing the flat-output $\bm{z}$, trajectory generation reduces to
constructing a sufficiently smooth curve in the flat-output space. We
represent this curve as a sequence of Hermite polynomial segments connecting flat-output waypoints \cite{kondo2026mighty}, where each segment prescribes derivatives through order $q$ at
its two endpoints. Adjacent segments agree through order $r$, the highest order entering the state recovery map $\Phi$, with $q > r$.

At each waypoint $w$ we store the flat-output together with its derivatives up
to order $r$,
\begin{equation}
    \mathcal{Z}_w = \bigl(\bm{z}_w, \bm{z}_w^{(1)}, \ldots, \bm{z}_w^{(r)}\bigr),
\end{equation}
and share this tuple between the two segments meeting at $w$. Since the state
recovery map $\Phi$ depends only on derivatives through order $r$, this keeps
the recovered state trajectory consistent across segments.

Derivatives of order $r+1,\ldots,q$ are set independently by each segment and
need not agree at the junction. These higher-order values do not enter $\Phi$ and
therefore act as free design variables that shape the segment without
affecting the continuity of the recovered state at junctions.

Each segment is parameterized by a normalized time $\sigma \in [0,1]$, with
$\sigma = 0$ at its start waypoint and $\sigma = 1$ at its end waypoint,
independent of the segment's duration in real time. This lets every segment
use the same fixed set of basis polynomials. For a segment connecting two adjacent waypoints $\bm{z}_w$ and $\bm{z}_{w+1}$, the Hermite curve is
\begin{equation}
    \bm{z}(\sigma)
    =
    \sum_{p=0}^{q}
    \Bigl[ h_p(\sigma)\, \bm{z}_w^{(p)} + g_p(\sigma)\, \bm{z}_{w+1}^{(p)} \Bigr],
    \label{eq:hermite}
\end{equation}
where $h_p, g_p$ are the degree-$(2q+1)$ Hermite basis polynomials
associated with the $p$-th derivative at the left and right endpoint, and the derivatives $\bm{z}_w^{(p)}$ are taken with respect to
$\sigma$. By construction,
\begin{equation} \label{eq:stitching}
    \bm{z}^{(p)}(0) = \bm{z}_w^{(p)},
    \qquad
    \bm{z}^{(p)}(1) = \bm{z}_{w+1}^{(p)},
    \qquad p = 0,\ldots,q,
\end{equation}
so adjacent segments share the tuple $\mathcal{Z}_w$ at every junction.

\begin{remark}
With $m$ flat-output components, $W$ waypoints, and $S = W-1$ segments, the
design vector holds $m(r+1)$ values per waypoint and $2m(q-r)$ per segment,
for a total dimension of $mW(r+1) + 2mS(q-r)$. This dimension depends on $W$
and $q$ alone and stays compact as the horizon grows.
\end{remark}

\begin{figure*}[t]
\centering
\includegraphics[width=\textwidth]{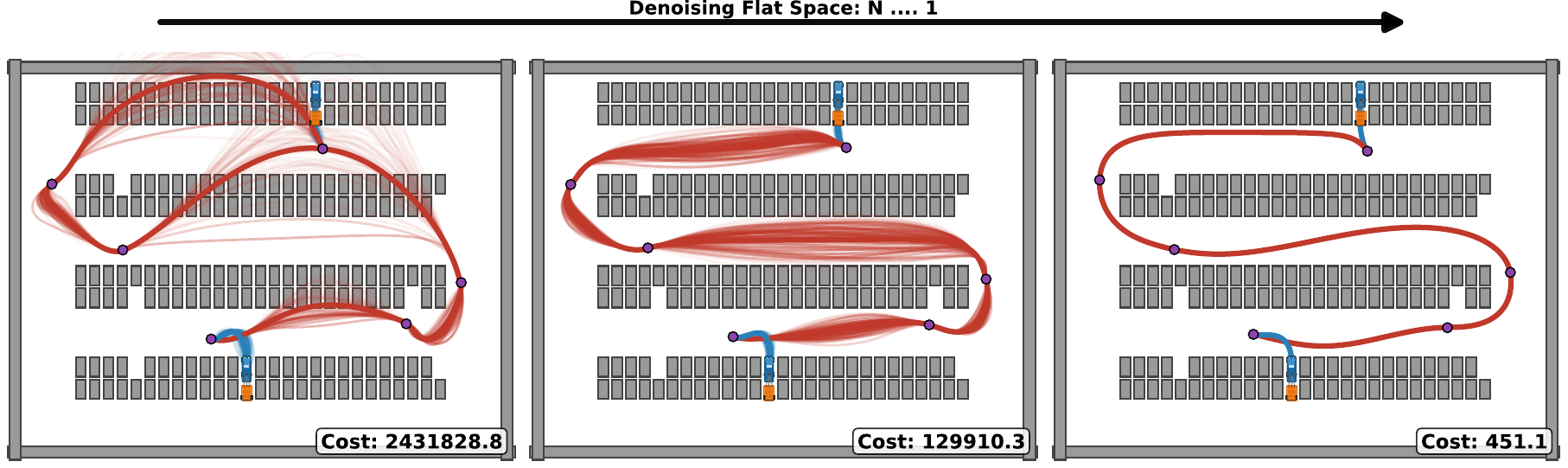}
\caption{Trajectory refinement during DFLAT denoising on a parking-lot scenario from the
\emph{Long} benchmark, shown at three stages from the coarse initialization
($i=N$) to the final solution ($i=1$). Red segments are forward maneuvers, blue
reverse, and purple markers the waypoints. Refining the waypoints and Hermite
parameters yields a smoother, lower-cost maneuver while remaining dynamically
consistent.}
\label{fig:diffusion_story}
\end{figure*}

\section{Diffusion on the Differentially Flat Space}
To exploit this compact representation, we formulate planning as a single
model-based diffusion process in the flat-output space, rather than diffusing
directly over $Y = (\tau_x, \tau_u)$. We target parking and maneuvering tasks
for nonholonomic articulated vehicles in cluttered environments, where reaching
the goal through non-convex free space may require several changes of driving
direction. A single smooth curve cannot express such a maneuver, so we
represent the flat-output by Hermite segments joined at sparsely placed
waypoints. The following proposition establishes that the concatenated
trajectory remains a solution of the dynamics, provided adjacent flat-output space segments agree
through order $r$.

\begin{proposition}
Let $\bm{z}$ concatenate finitely many Hermite segments agreeing through
order $r$ at every junction. Assume strictly increasing local timing maps
that are $C^{r+1}$ on each closed segment interval, with matching derivatives
of orders $1,\ldots,r$ at junctions. If the timed segments remain in the domain
of the smooth flatness maps $(\Phi,\Psi)$ and $\bm{u}(t)\in\mathcal{U}$,
then the recovered state $\bm{x}$ is absolutely continuous and satisfies
\begin{equation}
\dot{\bm{x}}(t)=f\bigl(\bm{x}(t),\bm{u}(t)\bigr)
\quad\text{for almost every }t\in[0,T].
\label{eq:stitched_dynamics}
\end{equation}
\end{proposition}

\begin{proof}
Let $\bar{\bm{z}}_s$ denote the timed segment on $[t_{s-1},t_s]$,
where $0=t_0<\cdots<t_S=T$. Each $\bar{\bm{z}}_s$ is $C^{r+1}$.
By the chain rule, matching normalized segment derivatives and timing
derivatives through order $r$ gives matching physical-time derivatives,
so the concatenated output $\bar{\bm{z}}$ is globally $C^r$.
On each open interval $(t_{s-1},t_s)$, the recovered pair
$\bm{x}=\Phi(\bar{\bm{z}}_s,\ldots,\bar{\bm{z}}_s^{(r)})$ and
$\bm{u}=\Psi(\bar{\bm{z}}_s,\ldots,\bar{\bm{z}}_s^{(r+1)})$
is well defined and satisfies \eqref{eq:stitched_dynamics}
by differential flatness. At every interior junction,
\begin{align}
\bm{x}(t_s^-)
&=\Phi\!\left(\bar{\bm{z}}_s(t_s),\ldots,
\bar{\bm{z}}_s^{(r)}(t_s)\right) \nonumber\\
&=\Phi\!\left(\bar{\bm{z}}_{s+1}(t_s),\ldots,
\bar{\bm{z}}_{s+1}^{(r)}(t_s)\right)=\bm{x}(t_s^+).
\end{align}
Thus $\bm{x}$ is continuous and, by smoothness of $\Phi$, $C^1$
on each closed segment interval, hence absolutely continuous.
Derivatives of order $r+1$ may differ at junctions, allowing jumps
in $\bm{u}$, while \eqref{eq:stitched_dynamics} holds almost everywhere.
\end{proof}

The proposition constrains only the derivatives shared at waypoints. Endpoint
derivatives of order above $r$ remain free, and we treat them as segment-local
degrees of freedom that reshape each segment without disrupting the endpoint
continuity between adjacent segments.
Each candidate pose $\bm{x}_w$ fixes the shared derivatives
$\bm{z}_w^{(0)},\ldots,\bm{z}_w^{(r)}$ through
$
\Phi\bigl(\bm{z}_w^{(0)},\ldots,\bm{z}_w^{(r)}\bigr)=\bm{x}_w,
$
while the remaining endpoint derivatives are denoted $\bm{z}_{s,\ell}^{(p)}$,
with $s$ the segment index, $\ell\in\{0,1\}$ the endpoint side, and
$p=r+1,\ldots,q$. The $W$ interior waypoints induce $W+1$ segments, and the
diffusion variable is
\begin{equation}
\bm{\eta}:=
\left(
\begin{aligned}
&\bm{x}_w, &&\! w=1,\ldots,W,\\[-1pt]
&\bm{z}_{s,\ell}^{(p)}, &&\! s=0,\ldots,W,\; \ell\in\{0,1\},\; p=r+1,\ldots,q
\end{aligned}
\right).
\label{eq:diffusion_variable}
\end{equation}

The start and goal states remain fixed throughout denoising. Each sample of
$\bm{\eta}$ defines a piecewise Hermite curve $\bm{z}(\cdot)$, which is lifted
through $(\Phi,\Psi)$ to obtain the state-input trajectory used to evaluate
$J$. Waypoint poses use smaller update magnitudes than the higher-order Hermite
derivatives, so the waypoint sequence settles early while the segments between
them stay free to refine. The process is initialized from a coarse
search-based prior in flat-output space, produced by Coarse Waypoint Planning
(CWP) in Algorithm~\ref{alg:cwp}, a cheaper variant of
Flat-RRT*~\cite{bascetta2017flat} that uses no precomputed motion primitives and does
not optimize solution quality.

CWP constructs this coarse initialization by growing a tree of poses in
flat-output space, where \textsc{Sample} draws state samples from a
neighborhood of an existing node and lifts it to the corresponding flat-output representation, \textsc{FlatConnect} generates a single Hermite connection
between two poses by fixing the flatness-relevant endpoint derivatives and heuristically initializing the remaining free derivatives, and
\textsc{Feasible} verifies the corresponding lifted trajectory against the map
$\mathcal{M}$. A connection is accepted if the lifted trajectory is collision-free and stays within a loosened bound on the articulation limits; the tighter state and input constraints are softly enforced during diffusion. The resulting pose sequence and associated Hermite parameters define the initial design $\bm{\eta}^{(M)}$.

\begin{algorithm}[h]
\caption{$\mathrm{CWP}$: Coarse Waypoint Planning}
\label{alg:cwp}
\begin{algorithmic}[1]
\Require map $\mathcal{M}$, poses $\bm{x}_{\mathrm{init}},\bm{x}_{\mathrm{goal}}$
\State $\mathcal{T}\!\gets\!\{\bm{x}_{\mathrm{init}}\}$
\Repeat
    \State $q_{\mathrm{rand}}\!\gets\!\mathrm{Sample}(\mathcal{M},\bm{x}_{\mathrm{goal}})$
    \State $q_{\mathrm{near}}\!\gets\!\mathrm{Nearest}(\mathcal{T},q_{\mathrm{rand}})$
    \State $e\!\gets\!\mathrm{FlatConnect}(q_{\mathrm{near}},q_{\mathrm{rand}})$
    \If{$\mathrm{Feasible}(e,\mathcal{M})$}
        \State add endpoint $q_{\mathrm{new}}$ of $e$ to $\mathcal{T}$
    \EndIf
\Until{$\mathrm{Feasible}(\mathrm{FlatConnect}(q_{\mathrm{new}},\bm{x}_{\mathrm{goal}}),\mathcal{M})$}
\State \Return $\bm{\eta}^{(N)}\!\gets\!\mathrm{Path}(\mathcal{T},\bm{x}_{\mathrm{init}},\bm{x}_{\mathrm{goal}})$
\end{algorithmic}
\end{algorithm}

Algorithm~\ref{alg:dflat} refines this initialization through model-based diffusion
directly on $\bm{\eta}$. At each reverse step, $K$ candidate designs
are sampled around the current estimate, lifted through the Hermite
construction and the flatness maps, and scored by the trajectory cost.
The cost-weighted average then determines the next denoising update,
following the model-based diffusion procedure of Section~\ref{sec:mbd} but applied
to $\bm{\eta}$ rather than $Y$.

\begin{algorithm}[h]
\caption{DFLAT}
\label{alg:dflat}
\begin{algorithmic}[1] 
\State $\bm{\eta}^{(N)} \gets \mathrm{CWP} (\mathcal{M},\bm{x}_{\mathrm{init}},\bm{x}_{\mathrm{goal}})$ \For{$i=N,\ldots,1$} 
\State $\{\bm{\eta}_k^{(i)}\}_{k=1}^{K} \gets \mathrm{Sample}(\bm{\eta}^{(i)})$ \For{$k=1,\ldots,K$} \State $\bm{z}_k \gets \mathrm{Hermite}(\bm{\eta}_k^{(i)})$ \Comment{fixed $\bm{z}^{(0:r)}$, free $\bm{z}^{(r+1:q)}$} 
\State $(\tau_{x,k},\tau_{u,k}) \gets (\Phi,\Psi)[\bm{z}_k]$ 
\State $J_k \gets J(\tau_{x,k},\tau_{u,k})$ \EndFor 
\State $\overline{\bm{\eta}}^{(i)} \gets \sum_k w_k \bm{\eta}_k^{(i)}$ where $w_k \propto e^{-J_k/\lambda}$ \Comment{Eq. \eqref{eq:softmax}}
\State $\bm{\eta}^{(i-1)} \gets \mathrm{ReverseStep} (\bm{\eta}^{(i)},\overline{\bm{\eta}}^{(i)}) $ \Comment{Eq. \eqref{eq:reversestep}}\EndFor 
\State $(\tau_x^*,\tau_u^*) \gets (\Phi,\Psi)\!\left[z^*(\bm{\eta}^{(0)})\right]$ 
\State \Return $(\tau_x^*,\tau_u^*)$ 
\end{algorithmic} 
\end{algorithm}

\section{Experiments}
\subsection{Dynamical Systems}
We evaluate DFLAT on a tractor-trailer (TT) system, whose kinematic model
has state $\bm{x} = [p^x,p^y,\theta^1,\theta^2]^\top$, where $(p^x,p^y)$
is the tractor position and $\theta^1,\theta^2$ are the tractor and
trailer heading angles, and control input $\bm{u}=[v,\delta]^\top$
(longitudinal velocity and steering angle):
\begin{align}
\dot{p}^x &= v\cos\theta^1,
&
\dot{\theta}^1 &= \frac{v}{\ell_1}\tan\delta, \\
\dot{p}^y &= v\sin\theta^1,
&
\dot{\theta}^2 &= \frac{v}{\ell_2}\sin(\theta^1-\theta^2),
\end{align}

\begin{figure}[h]
    \centering
    \includegraphics[width=0.8\columnwidth]{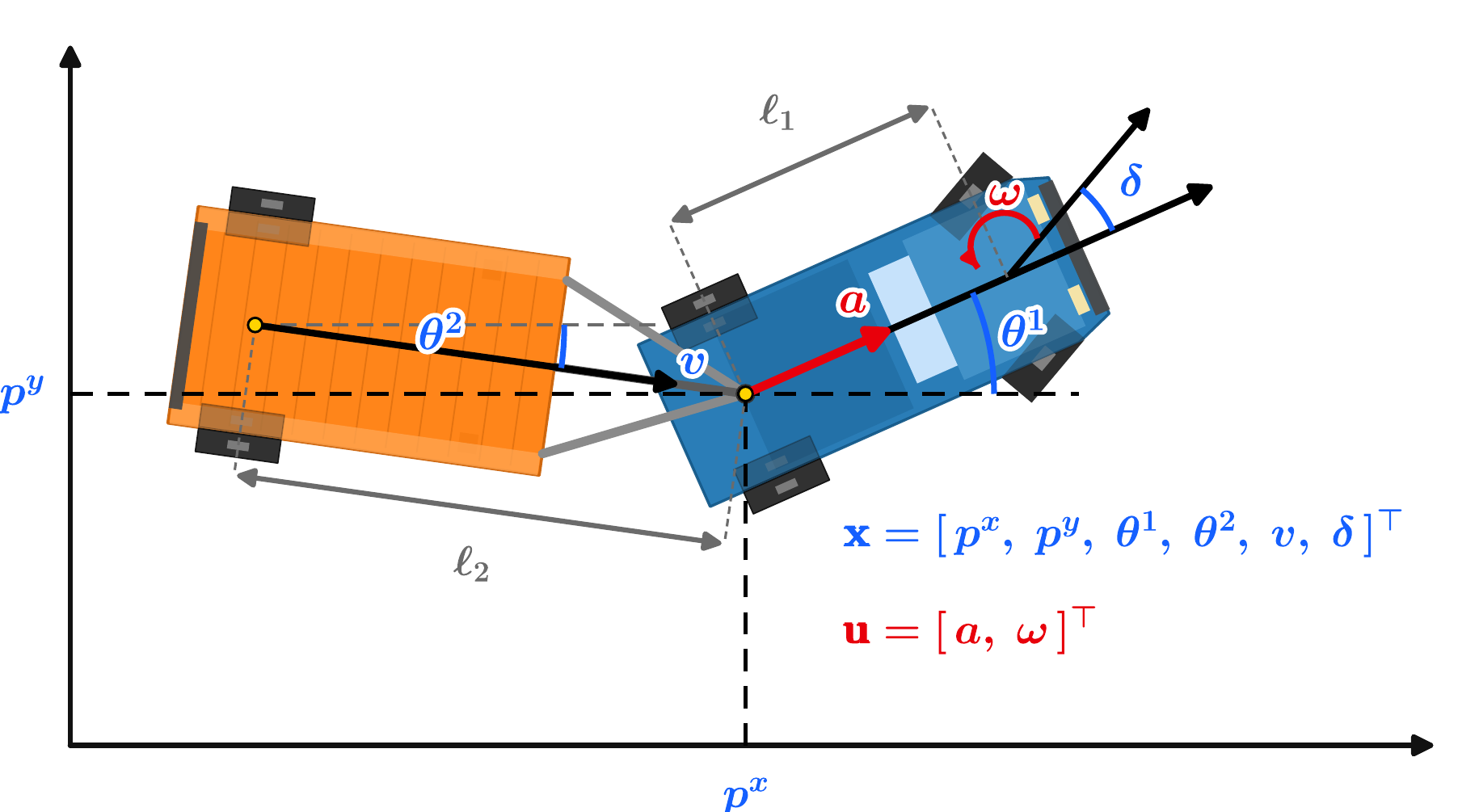}
    \caption{Acceleration-controlled tractor--trailer model
    (Accel-TT) used in the experiments, with acceleration $a$ and
    steering rate $\omega$ as control inputs.}
    \label{fig:tt_diagram}
\end{figure}

where $\ell_1,\ell_2$ are the tractor wheelbase and trailer length. We assume an on-axle hitch throughout, following~\cite{rouchon1993flatness};
off-axle configurations are outside the scope of this paper. We
additionally evaluate an acceleration-controlled variant
(Accel-TT), which augments the state with velocity and steering
angle, $\bm{x}_{\text{acc}}=[p^x,p^y,\theta^1,\theta^2,v,\delta]^\top$,
and takes $\bm{u}_{\text{acc}}=[a,\omega]^\top$ as input, adding
$
    \dot{v} = a, \quad \dot{\delta} = \omega.
$

\begin{table*}[t]
\centering
\scriptsize
\setlength{\tabcolsep}{4pt}
\renewcommand{\arraystretch}{0.9}
\caption{Performance comparison across difficulty levels and dynamics models. For each level, cases were aggregated across the constituent maps. Time is computed over solved runs only.}
\label{tab:benchmark}
\begin{tabular}{ll l cccc}
\toprule
\textbf{Level} & \textbf{Model} & \textbf{Metric} & \textbf{Kinodynamic RRT \cite{lavalle2001randomized}} & \textbf{Safe MPD \cite{kim2026safempd}} & \textbf{MPPI \cite{williams2015model}} & \cellcolor{gray!15}\textbf{DFLAT (Ours)} \\
\midrule
\multirow{4}{*}{\textbf{Short}}
 & \multirow{2}{*}{TT}
   & (A) Success (\%)  & \textbf{100.0\%} & 98.3\% & 86.7\% & \cellcolor{gray!15}\textbf{100.0\%} \\
 & & (B) Time [s] $\pm$ std  & 135.8$\pm$134.0 & 4.7$\pm$2.0 & 26.0$\pm$0.2 & \cellcolor{gray!15}\textbf{1.1$\pm$0.6} \\
\cmidrule(lr){2-7}
 & \multirow{2}{*}{Accel-TT}
   & (A) Success (\%)  & \textbf{100.0\%} & 68.3\% & 83.3\% & \cellcolor{gray!15}\textbf{100.0\%} \\
 & & (B) Time [s] $\pm$ std  & 119.3$\pm$204.7 & 7.6$\pm$3.1 & 26.2$\pm$0.2 & \cellcolor{gray!15}\textbf{1.2$\pm$0.7} \\
\midrule
\multirow{4}{*}{\textbf{Moderate}}
 & \multirow{2}{*}{TT}
   & (A) Success (\%)  & \textbf{100.0\%} & 33.3\% & 66.7\% & \cellcolor{gray!15}\textbf{100.0\%} \\
 & & (B) Time [s] $\pm$ std  & 210.5$\pm$394.0 & 6.2$\pm$0.4 & 25.9$\pm$0.1 & \cellcolor{gray!15}\textbf{1.5$\pm$0.7} \\
\cmidrule(lr){2-7}
 & \multirow{2}{*}{Accel-TT}
   & (A) Success (\%)  & \textbf{100.0\%} & 33.3\% & 90.0\% & \cellcolor{gray!15}\textbf{100.0\%} \\
 & & (B) Time [s] $\pm$ std  & 890.4$\pm$1757.1 & 9.9$\pm$0.4 & 26.1$\pm$0.1 & \cellcolor{gray!15}\textbf{1.6$\pm$0.7} \\
\midrule
\multirow{4}{*}{\textbf{Long}}
 & \multirow{2}{*}{TT}
   & (A) Success (\%)  & \textbf{100.0\%} & 5.0\% & 6.7\% & \cellcolor{gray!15}\textbf{100.0\%} \\
 & & (B) Time [s] $\pm$ std  & 182.4$\pm$217.7 & 43.2$\pm$1.0 & 59.3$\pm$0.1 & \cellcolor{gray!15}\textbf{3.1$\pm$2.1} \\
\cmidrule(lr){2-7}
 & \multirow{2}{*}{Accel-TT}
   & (A) Success (\%)  & \textbf{100.0\%} & 3.3\% & 8.3\% & \cellcolor{gray!15}\textbf{100.0\%} \\
 & & (B) Time [s] $\pm$ std  & 384.4$\pm$933.5 & 114.1$\pm$0.9 & 46.2$\pm$16.4 & \cellcolor{gray!15}\textbf{3.3$\pm$2.1} \\
\bottomrule
\end{tabular}
\end{table*}

\subsection{Reference Lifting and Tracking}

Each sampled design induces a piecewise flat-output curve that recovers,
through the flatness maps, a geometric path through the pose space
$(p^x,p^y,\theta^1,\theta^2)$. This path specifies the route geometry
but not its timing. Long-horizon maneuvers require the vehicle to stop
and reverse. At such a reversal the flat-output curve forms a cusp, where
the speed passes through zero and the direction of travel flips. Since
the headings are recovered from the direction of travel along the curve,
they are determined only where that direction is well defined, which
excludes the cusp itself. We therefore split the path at the reversals
into same-direction runs. Within a run the direction of travel never
changes, so the lift is well posed, and the speed profile is the fastest
traversal that respects the velocity, acceleration, and steering-rate
limits.

For TT, the resulting speed and steering are directly the control input
$\bm{u}$, whereas for Accel-TT they become components of the state
$\bm{x}_{\mathrm{acc}}$ and the inputs are their rates
$a=\dot v$ and $\omega=\dot\delta$. In both cases, the velocity and rate
bounds are satisfied by construction.  A downstream controller tracks this trajectory
during execution, as tractor--trailer systems are strongly unstable under
reverse motion~\cite{yang2025differential}.

\subsection{Experimental Setup}

\begin{figure*}[t]
\centering
\begin{minipage}[c]{0.42\textwidth}
  \centering
  \includegraphics[width=\linewidth]{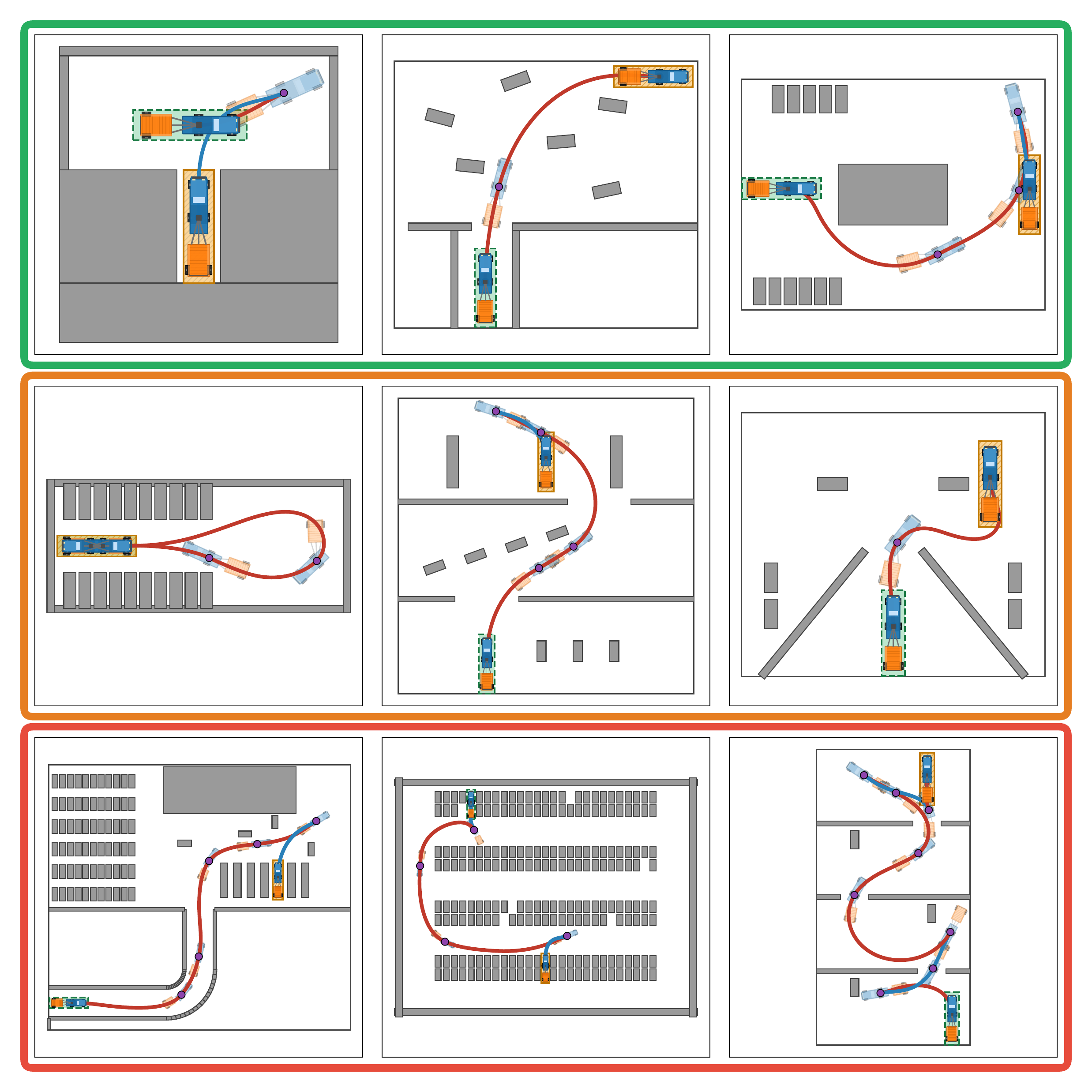}
\end{minipage}%
\hspace{2pt}%
\begin{minipage}[c]{0.50\textwidth}
  \centering
  \includegraphics[width=\linewidth]{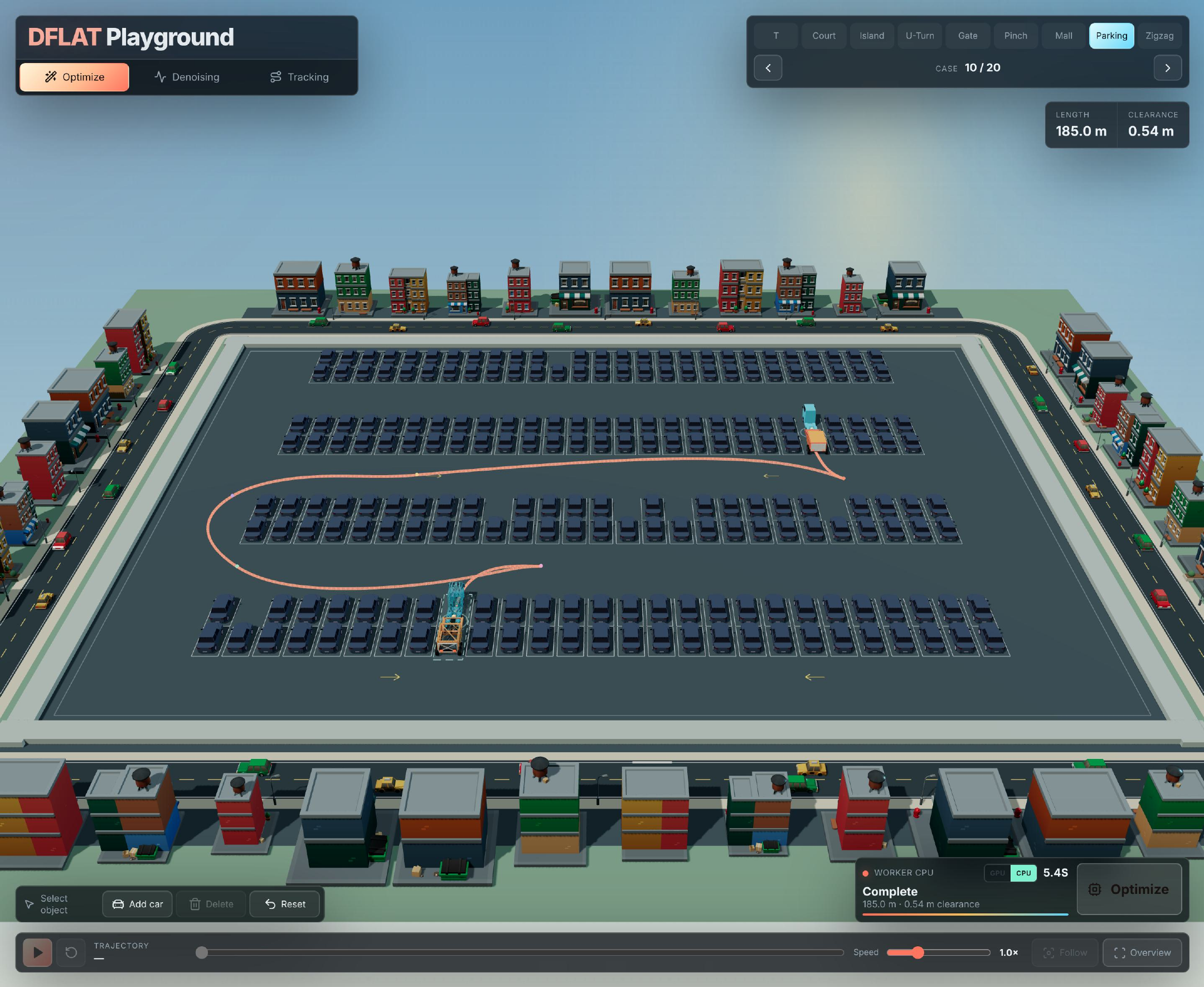}
\end{minipage}
\caption{Sample trajectories from DFLAT. Green slots mark the start pose, orange slots mark the goal. Red segments are forward maneuvers, blue segments are reverse maneuvers, and purple markers are the waypoints.
\protect\textbf{Left:} nine benchmark environments of increasing planning horizon.
\protect\emph{Top:} short (\protect\emph{T}, \protect\emph{Court}, \protect\emph{Island}).
\protect\emph{Middle:} medium, requiring a reversal or a tight passage
(\protect\emph{U-Turn}, \protect\emph{Compound Gate}, \protect\emph{Pinch Point}).
\protect\emph{Bottom:} long (\protect\emph{Shopping Mall}, \protect\emph{Parking Lot},
\protect\emph{Zig-Zag}).
DFLAT produces a collision-free trajectory respecting the dynamic limits in every case, at low latency.
\protect\textbf{Right:} \protect\emph{DFLAT Playground}, an interactive web-based demo in which all maps can be
tried with different start and goal poses and custom obstacle placement, with
DFLAT replanning in the browser.\protect\footnotemark{}}
\label{fig:maps_3by3_diffusion}
\end{figure*}

We evaluate DFLAT on a benchmark suite of nine parking and navigation maps
grouped into three difficulty levels: \emph{Short}, \emph{Moderate}, and \emph{Long}
(Fig.~\ref{fig:maps_3by3_diffusion}). \emph{Short} scenarios have relatively open
workspaces and brief maneuver sequences. \emph{Moderate} scenarios require tighter
maneuvers and, in some cases, globally coordinated ones such as U-turns and
gate traversals. \emph{Long} scenarios involve large-scale or densely cluttered
environments that demand long-horizon planning with multiple direction
changes. All experiments are run on a single NVIDIA RTX 5090 GPU.

Each difficulty level contains three scenario families, with 20 randomized
instances per family. Randomization is over the initial pose and, where the
map permits, the goal pose. Average trajectory length increases from roughly
20\,m in \emph{Short} scenarios to 150\,m in \emph{Long} ones, and the number of reversals
per instance ranges from 1 to 7. The three groups therefore correspond to
progressively longer planning horizons and increasing maneuver complexity.

DFLAT optimizes the objective
\begin{equation}
    J = J_{\mathrm{path}}
      + w_g J_{\mathrm{goal}}
      + w_k J_{\mathrm{kin}}
      + w_r J_{\mathrm{reg}}
      + w_c J_{\mathrm{coll}},
    \label{eq:cost}
\end{equation}
evaluated on the trajectory recovered from the flat-output trajectory. Here
$J_{\mathrm{path}}$ is the traversed path length and $J_{\mathrm{goal}}$ the
terminal pose error, with heading differences wrapped to $[-\pi,\pi]$. The
hitch angle is the relative orientation between tractor and trailer,
$
    \psi_t := \theta_{1,t} - \theta_{2,t}
$
and the trailer jackknifes when $|\psi_t| > \theta_{\max}$. The term
$J_{\mathrm{kin}}$ penalizes violation of the steering and hitch angle limits,
\begin{equation}
    J_{\mathrm{kin}} = \sum_{t}
    \max\bigl(0, |\delta_t| - \delta_{\max}\bigr)
    + \max\bigl(0, |\psi_t| - \theta_{\max}\bigr),
\end{equation}
while $J_{\mathrm{reg}} = \sum_t \delta_t^2 + \psi_t^2$
penalizes large steering and hitch angles, favoring
gentler maneuvers. Let $\mathcal{R}_{\mathrm{tr}}(\bm{x})$ and $\mathcal{R}_{\mathrm{tl}}(\bm{x})$
denote the tractor and trailer footprints at state $\bm{x}$, and let
$\bm{o}_j$ and $R_j$ denote the center and radius of obstacle $j$.
Collisions are penalized by
\begin{equation}
    J_{\mathrm{coll}} = \sum_{t} \sum_{j=1}^{N_{\mathrm{obs}}}
    \max\Bigl(0,\ R_j - \mathrm{dist}\bigl(
        \mathcal{R}_{\mathrm{tr}}(\bm{x}_t) \cup
        \mathcal{R}_{\mathrm{tl}}(\bm{x}_t),\ \mathcal{O}_j
    \bigr)\Bigr).
\end{equation}
All DFLAT experiments use sixth-order Hermite segments ($q = 6$), a 30-step
diffusion process, 192 samples per denoising step, and temperature
$\lambda = 0.1$. The same hyperparameters are used across all maps and both
dynamical systems, with no environment-specific tuning. Any missing parameters will be included with the code release.
A trial is successful if the tractor-trailer is fully contained in the goal
region $\mathcal{G}$ at the end of the executed trajectory
(see Fig.~\ref{fig:maps_3by3_diffusion}),
\begin{equation}
    \mathcal{R}_{\mathrm{tr}}(\bm{x}_T) \cup
    \mathcal{R}_{\mathrm{tl}}(\bm{x}_T) \subseteq \mathcal{G},
    \label{eq:success}
\end{equation}
and the trajectory is free of collision, jackknifing, and steering violation
within the planning budget, with thresholds $\delta_{\max} = 60^\circ$ and
$\theta_{\max} = 60^\circ$. The shooting-based methods, MPPI and Safe MPD,
roll out control sequences rather than optimizing over the terminal pose, so
we relax their goal condition to the tractor state lying inside $\mathcal{G}$.
This favors the baselines, as DFLAT is evaluated under a stricter, full-containment
condition.

\subsection{Main Benchmark Results}

We compare DFLAT against three planners for kinodynamic trajectory generation. Kinodynamic RRT \cite{lavalle2001randomized} serves as a feasibility reference rather than a competing planner, MPPI \cite{williams2015model} a standard shooting based method that rolls out perturbed control sequences and combines them by exponential cost weighting, and Safe Model Predictive Diffusion (Safe MPD) \cite{kim2026safempd} is a recent diffusion planner that uses shielded rollouts to keep sampled trajectories collision-free, evaluated on tractor-trailer systems.

Table~\ref{tab:benchmark} summarizes the results across all difficulty levels
and both dynamics models. Safe MPD is evaluated with a planning horizon of 100
steps, 20{,}000 samples per denoising step, and a 120-step denoising process;
MPPI uses the same horizon and sample count over 120 iterations. Since these
shooting-based planners are sensitive to hyperparameters, we tuned them with
Optuna, using a single setting across all maps. Among the baselines, Kinodynamic
RRT solves every instance, but at substantially higher computation time. Its
planning time is heavy-tailed, with some runs taking far longer than others; we
report the standard deviation for consistency with the other methods. DFLAT achieves the
highest overall success rate, reaching 100\% on all TT and Accel-TT
benchmarks, and plans faster than MPPI and Safe MPD on every benchmark. Safe MPD reaches 98\% success on \emph{Short} scenarios with TT but drops to 69\% under Accel-TT, while MPPI holds 83--87\% across both. Both methods
degrade on \emph{Moderate} scenarios, falling as low as 33.3\%, and fail almost
entirely on \emph{Long} instances, where the best success rate is 8.3\%. Their failure modes
differ: Safe MPD mostly avoids constraint violations but is often conservative and stops short of the goal, whereas MPPI can reach the goal more frequently and its failures generally come from collision. These results indicate that DFLAT remains effective on the long-horizon and constrained scenarios where control-space methods struggle, without giving
up dynamic consistency.

\subsection{Isolating the Representation from the Optimizer}
The results above combine the effects of the trajectory representation, the search-based initialization, and the diffusion optimizer. We next isolate
these factors by comparing planners that share the same differentially flat
representation. We evaluate these additional experiments only for the \emph{Long}
planning problems with the Accel-TT dynamics.

\begin{figure}[h]
    \centering
    \includegraphics[width=.84\columnwidth]{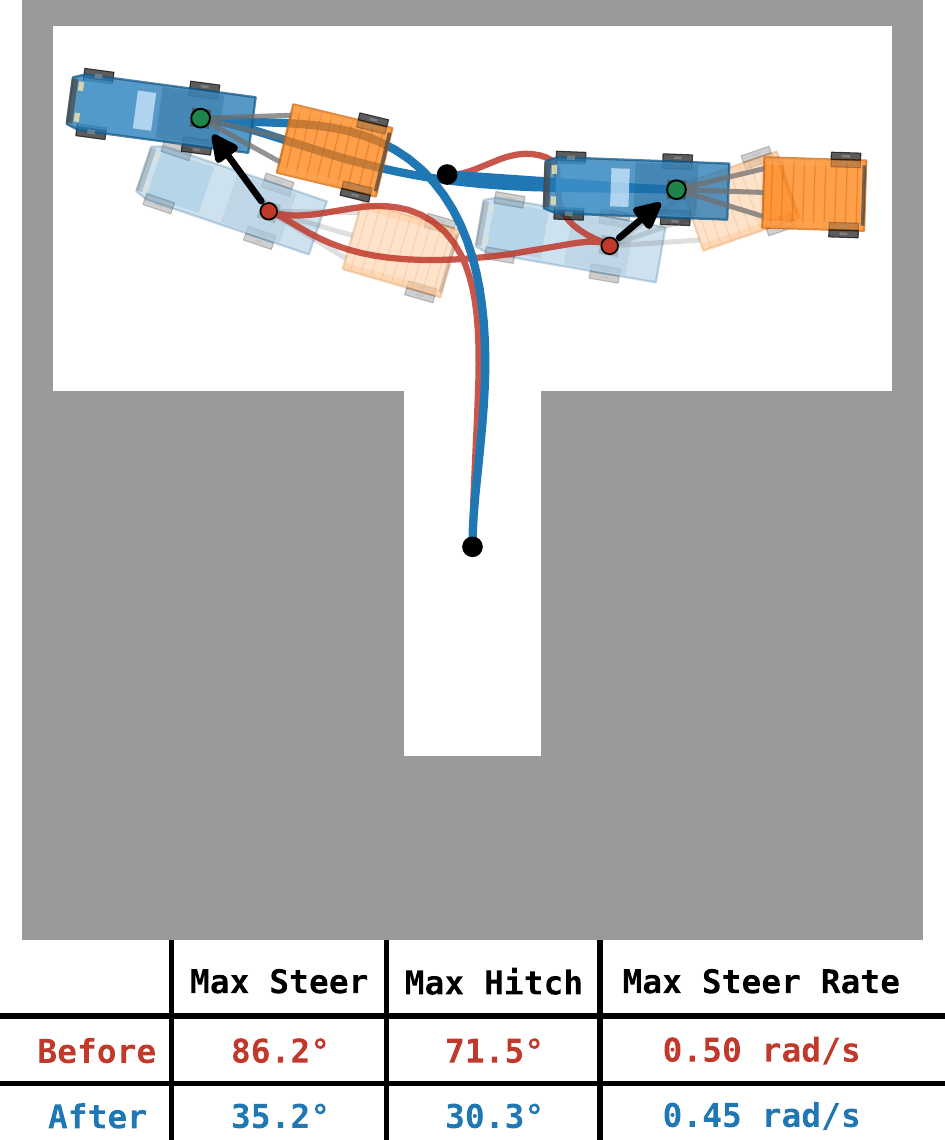}
    \caption{CWP-only trajectory (red) and full DFLAT trajectory (blue). CWP cheaply recovers a collision-free but dynamically infeasible reference. DFLAT jointly optimizes waypoints and segment derivatives, yielding a smooth, feasible trajectory.}
    \label{fig:t_waypoints_before_after}
\end{figure}

We first examine whether the search-based front end alone suffices, that is,
whether running the search to a higher-quality solution removes the need for a
refinement stage. Kinodynamic RRT, reported separately in
Table~\ref{tab:benchmark}, is the most direct version of this and requires
extremely large planning times, and tightening the feasibility tolerance of CWP
is likewise slow, as the resulting sample rejection rate dominates the runtime.
We therefore compare against Flat-RRT*~\cite{bascetta2017flat}, which plans over
a library of precomputed motion primitives in flat-output space. CWP is a
stripped-down version of the same search without precomputed primitives,
producing a feasible but low-quality trajectory that DFLAT uses as its prior;
Figure~\ref{fig:t_waypoints_before_after} shows the quality difference between
the CWP-only trajectory and the DFLAT output.

\footnotetext{\href{https://dflat.taekyung.me/}{\textcolor{black}{\nolinkurl{https://dflat.taekyung.me/}}}}

We then assess whether the diffusion optimizer is necessary once such a prior
is available. We first compare with CWP+GS, which refines the same prior by
perturbing the design with a fixed Gaussian and keeping the best candidate,
using the same number of samples and iterations as DFLAT. We further compare
with CWP+L-BFGS, which refines the prior with gradient-based L-BFGS
optimization \cite{liu1989lbfgs} over 30 gradient steps
on the same cost function, isolating the contribution of the diffusion stage
from that of the initialization.

\begin{table}[h]
\centering
\small
\setlength{\tabcolsep}{3pt}
\renewcommand{\arraystretch}{0.95}
\caption{Comparison on long-horizon scenarios under a common flat-output representation.}
\label{tab:ablation}
\begin{tabular}{lcccc}
\toprule
Metric & CWP+GS & CWP+L-BFGS & Flat-RRT* & \cellcolor{gray!15}DFLAT \\
\midrule
Success (\%) & 85.0 & 96.7 & \textbf{100.0} & \cellcolor{gray!15}\textbf{100.0} \\
Time (s) & \textbf{3.5$\pm$2.3} & 4.8$\pm$2.3 & 26.5$\pm$33.1 & \cellcolor{gray!15}\textbf{3.3$\pm$2.1} \\
Steer (deg) & 40.3 & 44.4 & 45.4 & \cellcolor{gray!15}\textbf{35.9} \\
Hitch (deg) & 30.7 & 33.4 & 31.5 & \cellcolor{gray!15}\textbf{28.6} \\
Rate (rad/s) & \textbf{0.440} & \textbf{0.453} & 0.481 & \cellcolor{gray!15}\textbf{0.456} \\
\bottomrule
\end{tabular}
\end{table}

Table~\ref{tab:ablation} reports the results, where the steering, hitch, and
steering-rate rows report the maximum value observed along the trajectory, and
values within $1\%$, $1^\circ$, $0.2$~s, or $0.2$~rad/s of the best are also
bolded. DFLAT attains the highest success rate at lower planning time than
either alternative, and produces trajectories with lower peak steering and
hitch angles. Flat-RRT* also reaches a perfect success rate, but takes
substantially longer to produce a trajectory of comparable quality, which is
why DFLAT initializes from the much cheaper CWP prior instead.

\section{Conclusion}
We presented DFLAT, a diffusion-based trajectory optimization framework that
operates directly in a differentially flat representation. Diffusing over
Hermite-curve parameters in the flat-output space preserves dynamic consistency throughout optimization and avoids the horizon-dependent
control-sequence representation used by existing diffusion planners. The
representation is compact enough to optimize the waypoint sequence and segment
geometry in a single diffusion process, rather than separating subgoal
planning from trajectory generation as hierarchical methods do. On
long-horizon tractor--trailer problems, DFLAT matches or exceeds the success
rates of the baselines at one to two orders of magnitude lower planning time.

The approach is not specific to tractor--trailer systems. The same
construction applies in principle to any differentially flat system, including
quadrotors or fixed-wing aircraft, enabling fast diffusion-based planning for a
broad class of nonholonomic dynamics. Natural extensions are to replace the CWP
prior with a learned diffusion model over the initial waypoints, and to let the
number of segments be discovered during denoising rather than fixed in advance.

\bibliographystyle{IEEEtran}
\bibliography{software}

\end{document}